\documentclass[runningheads]{llncs}

\usepackage[T1]{fontenc}
\usepackage{caption}
\usepackage{cite}
\usepackage{amsmath,amssymb,amsfonts}
\usepackage{algorithmic}
\usepackage{enumitem}
\setlist{nosep}
\usepackage{subcaption}
\usepackage{tabularx}
\usepackage{textcomp}
\usepackage{adjustbox}
\usepackage{booktabs}
\usepackage{float}
\usepackage{graphicx}
\usepackage{xcolor}
\usepackage{hyperref}

\begin{document}

\title{Using Poly-Encoders for Computationally Efficient Automated Creativity Assessment}

\titlerunning{Poly-Encoders for Creativity Assessment}

\author{Sam Grouchnikov\orcidID{0009-0007-5327-002X}\inst{1} \and 
Phillip Gregory\orcidID{0009-0004-1920-7607}\inst{2} \and 
Jiho Noh\orcidID{0000-0001-5734-9068}\inst{2}}

\authorrunning{S. Grouchnikov et al.}

\institute{Wheeler Magnet High School, Marietta, Georgia, USA\\
\email{sam.grouchnikov@gmail.com} \and
Kennesaw State University, Marietta, Georgia, USA\\
\email{pgrego10@students.kennesaw.edu, jnoh3@kennesaw.edu}}

\maketitle         

\begin{abstract}
Automated creativity assessment has been a long standing challenge, with traditional methods often being resource intensive or lacking practical accuracy. We introduce a novel approach by using Poly-Encoder for computationally efficient and accurate automated creativity assessment. We fine-tuned a Poly-Encoder on a public dataset from the Scientific Creative Thinking Test, comprised of approximately 18,000 human-rated question responses. Our method leverages small pre-trained BERT encoders, achieving performance comparable to fine-tuned Large Language Models while significantly reducing computational demands. Experiments with the BERT-family models and poly-code counts achieved Pearson 
correlations of up to $r = 0.74, 95\% \text{ CI } [0.73, 0.75]$ 
with human raters, matching the performance of resource intensive LLMs. This study bridges the gap between high performance and computational efficiency, potentially enabling widespread implementation of automated creativity assessment on accessible consumer-grade hardware. With some limitations, our findings suggest that Poly-Encoders are a promising alternative to LLMs for practical, scalable creativity assessment in various contexts, especially educational.

\end{abstract}

\section{Introduction}
Creativity is the basis of human progress, enabling the generation of novel, valuable, and feasible ideas across domains such as science, technology, engineering, and mathematics. Yet, assessing creativity remains a persistent challenge, highlighted best by creativity’s subjective nature \cite{Katz1982-de}. Traditional approaches rely on human judges, proving the process to be both time and resource-intensive \cite{Dumas2020-qj, Acar2024-nw, Haase2025-nn}. Even with shared definitions and common rubrics, inconsistency often emerges between evaluators \cite{Dumas2020-qj, Cropley2012-fr, Barth2021-fg}. This raises the central question: how can artificial intelligence be used to automate creativity assessment while still capturing the human creative intent?

Effective AI-based creativity assessment requires managing test complexity, efficient deployment, and alignment with human intent \cite{Rafner2023-rq}. Transformers are ideal for this task as they are designed to accurately contextualize text, and capture complex semantic nuances \cite{Hill2024-yu}. Through fine-tuning, such transformers may be trained to identify linguistic patterns associated with various types of creativity \cite{Bellemare-Pepin2025-nr, Organisciak2023-cw}.

In scientific contexts, creativity involves a \textit{dual-space search}: generating hypotheses from memory and solving problems by generalizing from experiments \cite{Beaty2024-kz, Klahr1988-qk}. Automated models must therefore understand scientific reasoning while aligning with psychometric criteria including fluency, originality, and elaboration \cite{Glover1976-ld, Handayani2021-du}. This is particularly important in education, where creativity-based assessments push beyond strict memorization to test divergent thinking and problem-solving \cite{Beghetto2010-cf, Lucas2013-sb}.

However, traditional human evaluation scales poorly. While Large Language Models (LLMs) offer a solution, they can be unreliable and sensitive to small input changes \cite{DhinakaranUnknown-vs}. Furthermore, reliance on cloud-based LLMs like GPT-4 introduces significant data privacy risks and recurring API costs that may be prohibitive for school districts. In contrast, local deployment ensures that sensitive student data remains within the school’s secure infrastructure. Efficient alternatives like Poly-Encoders \cite{Humeau2019-gy} provide a more reliable, scalable, and accessible path for real-time assessments using consumer-grade hardware available in classrooms. 

In this study, the implementation and analysis of the Poly-Encoder \cite{Humeau2019-gy} is used to answer the following questions:
\begin{enumerate}
    \item How can a Poly-Encoder be trained to assess creativity instead of similarity?
    \item How computationally efficient are Poly-Encoders, and can they be used on consumer hardware?
\end{enumerate}

\section{Related Work}
\subsection{Evolution of Creativity Assessment}
Early automated assessments relied on semantic distance models such as SemDis and GloVe. While effective at measuring semantic (dis)similarity through word-level relationships, they fail to capture semantic nuance and sentence-level contexts, both of which are essential for creativity assessment \cite{Organisciak2023-cw}. This resulted in a shift towards the utilization of Large Language Models (LLMs) that embed meaning at the sentence level and beyond. Initial prompt-based methods using GPT-4 achieved Pearson correlations between 0.2 and 0.67 with human ratings for novelty and feasibility \cite{Kern2024-hx}. However, these systems struggle with absolute scoring, and lack reliability due to their reliance on calling external APIs \cite{Rabeyah2024-pu}.

\subsection{Fine-Tuning Pre-Trained Language Models}
The next wave of methods used fine-tuning, training language models on questions with responses scored by human raters. Fine-tuning GPT-3 and T5 on 27,000 alternative uses task responses yielded correlations of up to $r = 0.81$  \cite{Organisciak2023-cw}. Similarly, the Scientific Creative Thinking Test (SCTT) utilized a dataset of 18,000 responses to fine-tune LLaMA-2-7b, reaching a Pearson correlation of $r=0.74$ \cite{Beaty2024-kz}. Although these methods exhibited strong scoring capability, their dependence on LLMs and the requirement for re-encoding questions/responses makes them computationally sub-optimal

\subsection{The Poly-Encoder Alternative}
Poly-Encoders \cite{Humeau2019-gy} provide a middle ground between the efficiency and semantic distance and the accuracy of fine-tuned LLMs. Poly-Encoders project context (or query) embeddings into multiple spaces by using a fixed-size set of learnable global attention codes \cite{Bahdanau2014-ly, VaswaniUnknown-ut}, providing potential to learn deep semantic representations of text and different facets of creativity without the computational overhead of LLMs. When paired with lightweight encoders like BERT \cite{Devlin2018-ss}, poly-encoders deliver consistent, reliable scoring that can be deployed on consumer-grade hardware: a task impossible with LLMs. 

\section{Methods}
Due to the potential applications and implications of accurate automated creativity assessment within educational and professional environments, a Poly-Encoder was adopted and trained to evaluate responses to questions. These questions were specifically centered around scientific prompts, including research inquiries and hypothesis formulation. The Poly-Encoder was implemented utilizing PyTorch Lighting, and training was executed in a distributed data parallel (DDP) configuration across three Nvidia RTX 3090 GPUs.

\subsection{Datasets and Training Schematics}
All training and testing data were obtained from the public repository released by the authors of the Scientific Creativity Thinking Task (SCTT) validation study~\footnote{\url{https://osf.io/preprints/psyarxiv/y5fbs_v1}}. The dataset contained approximately 18,000 responses, each paired with human-assigned ground-truth creativity labels corresponding to specific prompts. There are 15 prompts, each with approximately 1,200 responses. In our system, the prompt is the input to the Poly-Encoder as the context, while the response is the input as the candidate (see Figure~\ref{fig:combined_polyencoder}). To confirm the validity and reliability of the SCTT test, the authors employed multiple statistical tests. Inter-rater reliability achieved a coefficient of \(w = 0.85\) and a test-retest analysis yielded a moderate temporal stability with a Pearson correlation of $r=0.67$ across a one month interval. A conventional 70/10/20 train/validation/test split was employed (at the response level). Prompts in the training set were also included in the test set, while responses in the training set were not used in the test set. Within prompts, 70\% of responses were allocated to the training set, 10\% to validation, and 20\% to testing.

\subsubsection{BERT for Context/Candidate Embeddings}
Bidirectional Encoder Representations from Transformers (BERT) serves as a robust foundation for semantic representation learning \cite{Devlin2018-ss}, giving a favorable combination of learning potential and computational efficiency. To capture nuanced representations of creativity, multiple BERT-family models were explored for context and candidate encodings, gradually scaling in size to investigate performance trade-offs (Table~\ref{tab:bert variations}).

\begin{table}[ht!]
  \setlength{\tabcolsep}{8pt}
  \centering
  \caption{Tested BERT Models with their Sizes and Parameter Counts}
  \label{tab:bert variations}
    \begin{tabular}{@{}lrrr@{}}    \toprule
    \textbf{BERT Variation} & \textbf{Hidden Size} & \textbf{\# Layers} & \textbf{Parameter Count} \\
    \midrule
    BERT-Base & 768 & 12 & 110M \\
    BERT-Large & 1,024 & 24 & 340M \\
    RoBERTa-Base & 768 & 12 & 125M \\
    RoBERTa-Large & 1,024 & 24 & 355M \\
    DeBERTa-Base & 768 & 12 & 183M \\
    DeBERTa-v3-Large & 1,024 & 24 & 435M \\
    \bottomrule
  \end{tabular}
\end{table}

\subsubsection{Testing Poly-Code Counts} As described in the original Poly-Encoder paper, varying the number of poly-codes (\(m\)) can influence performance and computational cost. Ideally, reducing \(m\) should not substantially degrade results, indicating efficiency in representation. All hyper-parameters aside from poly-code count were held constant during testing. Tested poly-code counts were 64, 128, 256, and 512.

\subsection{Architectural Tweaks}
\begin{figure}[t]
     \centering
     \begin{subfigure}[b]{0.48\textwidth}
         \centering
         \includegraphics[width=\textwidth]{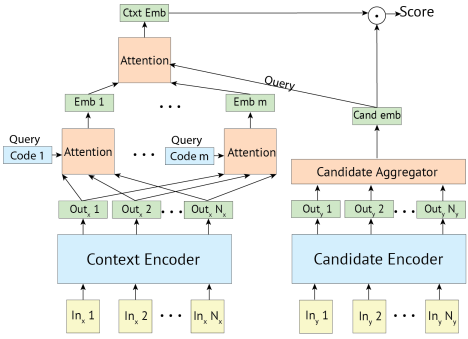}
         \caption{Original Poly-Encoder.}
         \label{fig:polyencoderarch}
     \end{subfigure}
     \hfill %
     \begin{subfigure}[b]{0.48\textwidth}
         \centering
         \includegraphics[width=\textwidth]{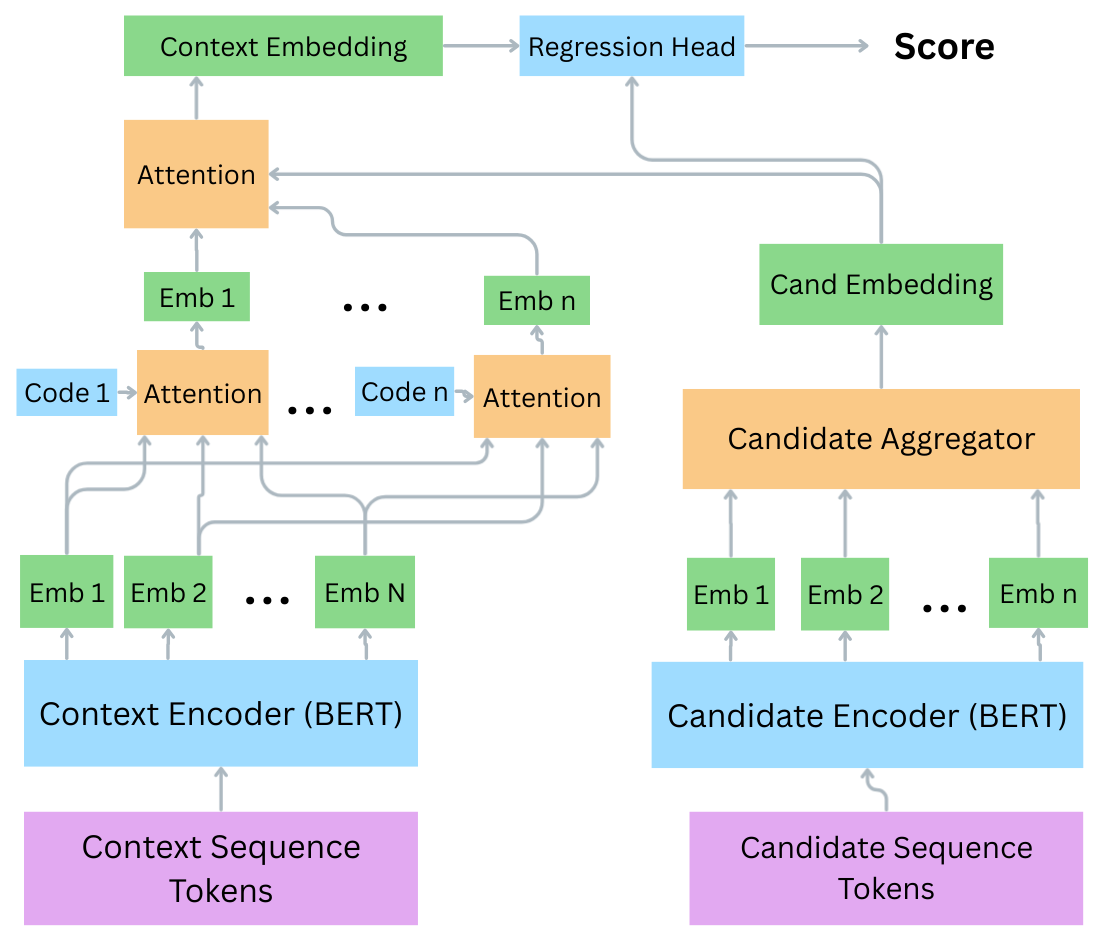}
         \caption{Modified Poly-Encoder.}
         \label{fig:polyencodermodified}
     \end{subfigure}
     
     \caption{Comparison of architectures (original vs our modified version)}
     \label{fig:combined_polyencoder}
\end{figure}

The base architecture of the Poly-Encoder was adopted from the paper released by Humeau et al. \cite{Humeau2019-gy}. The flow of information and calculations are depicted in Figure~\ref{fig:polyencoderarch} 
The original Poly-Encoder architecture produces a similarity score by computing the dot product between the poly-context and the candidate embeddings. While effective for retrieval and matching tasks, this mechanism underperformed for deeper scoring tasks, such as in our case, creativity. To address this, the final score was obtained by passing the poly-context embeddings through a lightweight regression head, allowing the model to recognize non-linearities of the final embeddings. This head also processed candidate vectors and the element-wise products of poly-context and candidate vectors, addressing the model's initial difficulty in differentiating between candidates. The regression head itself consisted of a multiple linear layers, along with an activation function and dropout. This architectural adjustment is visually illustrated in Figure~\ref{fig:polyencodermodified}.

\section{Results}

\subsection{BERT Comparisons} 

Progressively larger BERT models were tested while keeping all other aspects of the model constant. To ensure correct convergence, a plot was obtained from the Weights \& Biases logger. These plots are shown in Figure~\ref{fig:bert-test-corr}. The final highest correlations achieved by the BERT models are shown in Table~\ref{tab:db_v_rb_polym}. DeBERTa-v3-Large achieved the strongest correlation with human creativity scores $(p<0.001)$, outperforming all other tested encoders. This result aligns with expectations given DeBERTa’s disentangled attention mechanism \cite{He2020-nv}, which models content and positional information through separate vectors. However, RoBERTa variants demonstrated comparable results, suggesting that smaller, computationally efficient models may achieve similar accuracy. It is important to note however, all of the poly-code counts produced $p<0.05$ between DeBERTa and RoBERTa variants.

\subsection{Poly-Code Count Comparisons} 

\begin{figure}[t]
    \centering
    \begin{subfigure}[b]{0.48\textwidth}
        \centering
        \includegraphics[width=\textwidth]{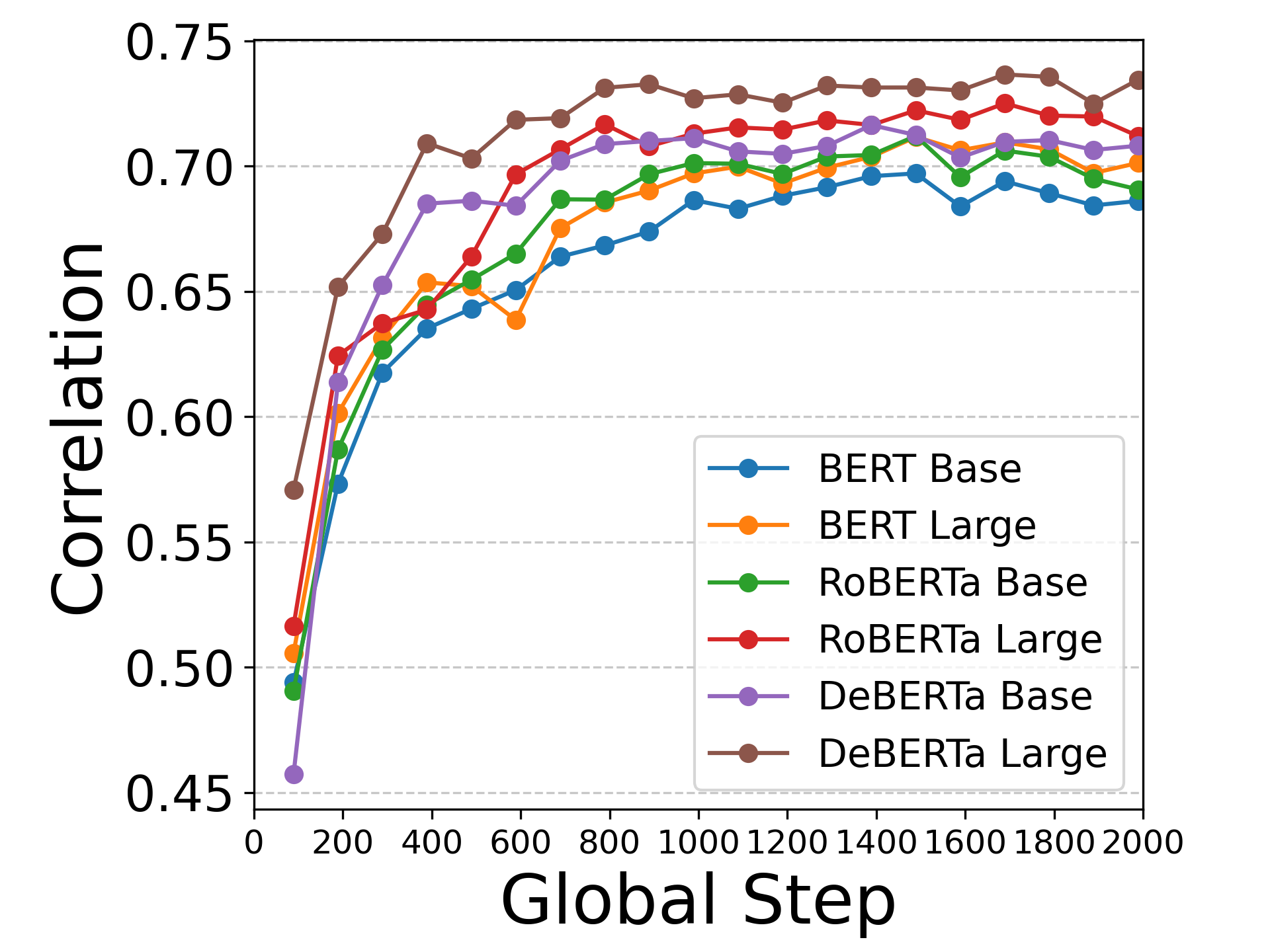}
        \caption{BERT Model Comparison}
        \label{fig:bert-test-corr}
    \end{subfigure}
    \hfill
    \begin{subfigure}[b]{0.48\textwidth}
        \centering
        \includegraphics[width=\textwidth]{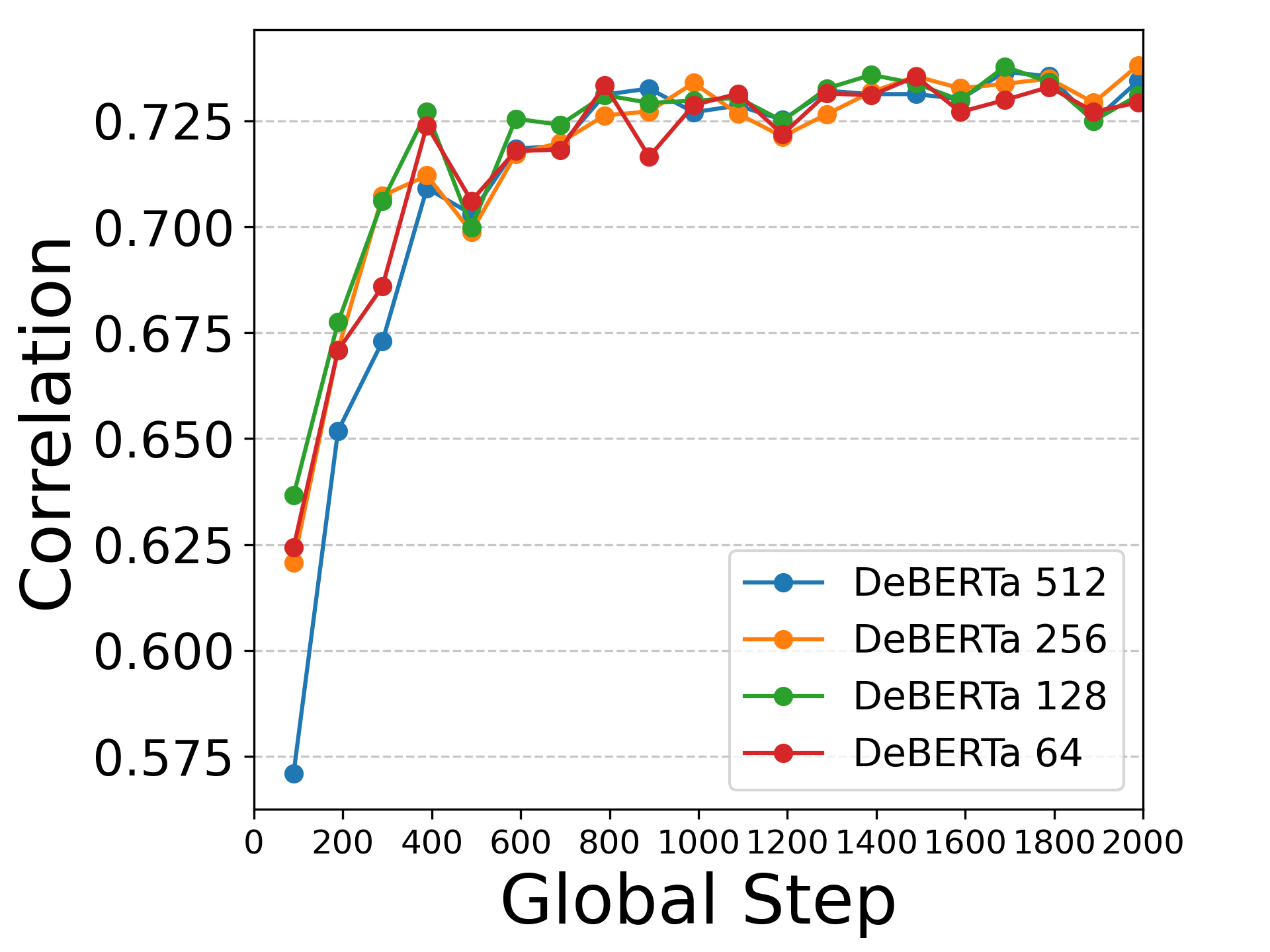}
        \caption{Poly-code ($m$) Comparison}
        \label{fig:deberta-test-corr}
    \end{subfigure}

    \caption{Training convergence and Pearson correlation results}
    \label{fig:training_comparisons}
\end{figure}

Using DeBERTa-v3-Large as the encoder, poly-code counts of 64, 128, 256, and 512 were evaluated, with test correlation plots shown in Figure~\ref{fig:deberta-test-corr}. Additionally, RoBERTa-Base was chosen for poly-code count comparisons due to its desirable mix of low parameter count and strong correlation. Comparisons are shown in Table~\ref{tab:db_v_rb_polym}. It should be noted, DeBERTa-v3-Large in a Poly-Encoder achieved the same correlation as LLaMA-2-7B~\cite{Beaty2024-kz} while using a fraction of the resources as LLaMA.

\begin{table}[ht!]
  \centering
  \caption{Performance comparison across poly-code counts ($m$) with 95\% CI}
  \label{tab:db_v_rb_polym}
  \begin{tabular}{l l l}
    \toprule
    \textbf{Poly-Codes ($m$)  } & \textbf{DeBERTa-v3-Large ($r$)  } & \textbf{RoBERTa-Base ($r$)} \\
    \midrule
    64  & 0.73 [0.72, 0.74] & 0.71 [0.70, 0.72] \\
    128 & 0.74 [0.73, 0.75] & 0.72 [0.71, 0.73] \\
    256 & 0.73 [0.72, 0.74] & 0.72 [0.71, 0.73] \\
    512 & 0.74 [0.73, 0.75] & 0.70 [0.69, 0.71] \\
    \bottomrule
  \end{tabular}
\end{table}

\begin{figure}[t]
    \centering
    \begin{subfigure}[b]{0.4\textwidth}
        \centering
        \includegraphics[width=\textwidth]{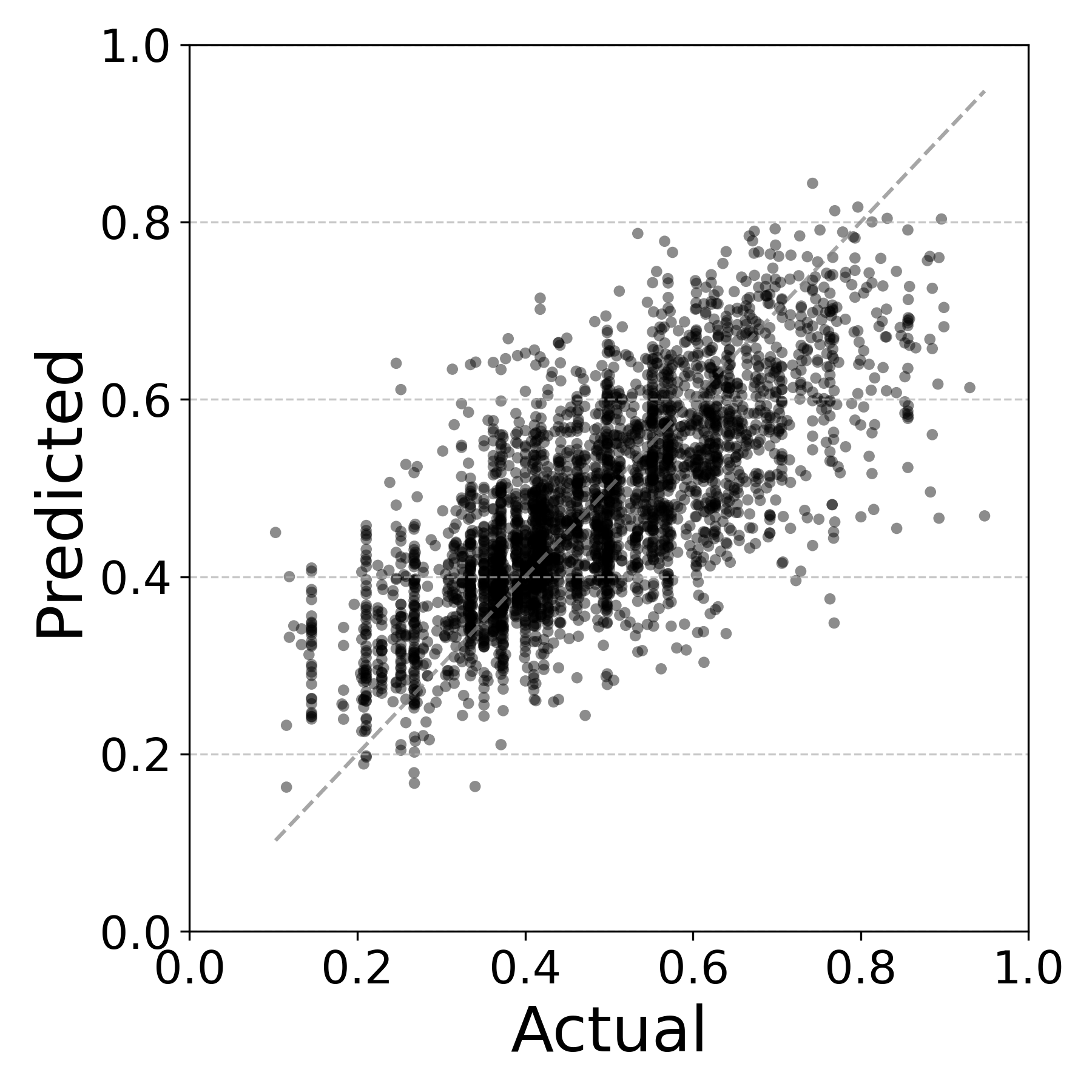}
        \caption{Human vs Model Scores}
        \label{fig:final-scatterplot}
    \end{subfigure}
    \hfill
    \begin{subfigure}[b]{0.58\textwidth}
        \centering
        \begin{tabular}[b]{@{}rr@{}}
            \toprule
            \textbf{\% Threshold} & \textbf{\% of Responses} \\
            \midrule
            1  & 4.0  \\
            5  & 20.9 \\
            10 & 40.1 \\
            15 & 56.0 \\
            25 & 78.5 \\
            50 & 95.2 \\
            \bottomrule
        \end{tabular}
        \vspace{1.5em} 
        \caption{Model Scoring Threshold Accuracy}
        \label{tab:model_threshold_accuracy}
    \end{subfigure}

    \caption{Performance Analysis: (a) visualizes the correlation between predicted and ground-truth scores, while (b) provides the exact cumulative accuracy at various error thresholds.}
    \label{fig:model_performance_combined}
\end{figure}

The model's predicted scores versus human ground-truth ratings are shown in Figure~\ref{fig:final-scatterplot}. The model demonstrated strong alignment with human ratings, correctly predicting the majority of creativity scores within reasonable threshhold (\% difference between ground-truth and predicted scores), as shown in  Table~\ref{tab:model_threshold_accuracy}. Additionally, with DeBERTa-v3-Large, the mean absolute error (MAE) between ground-truth and predicted scores was 0.0077, further proving the accuracy of the model.

\subsection{Efficiency, Scoring Times}
Although not performing better in terms of correlation from previous approaches, the true benefit of Poly-Encoders comes in their efficiency. The final scoring times (per candidate) for the DeBERTa-v3-Large and RoBERTa-Model were 0.010 and 0.0022 seconds on a CPU (Intel i5, common consumer-grade CPU) respectively.

\section{Conclusions}

\textbf{Performance and Efficiency}:  While Poly-Encoders do not outperform large language models in terms of absolute correlation, they achieve comparable performance with significantly higher computational efficiency. By encoding prompts once and utilizing lightweight BERT-based encoders, Poly-Encoders bypass the massive memory requirements presented by large models such as GPT-3 and LLaMA-2. This architecture allows for sub-second scoring on consumer grade hardware: a feature necessary for real-world deployment and use in classrooms with limited resources.

\textbf{Poly-Code Impact}: Consistent with existing literature, variations in the number of poly-codes ($m$) had no significant effect on model performance. This confirms that poly-codes enhance representational power while having a negligible effect on model size. The only significant affecting factor of performance was the family and size of the BERT encoder used ($p<0.05$ between BERT families with same poly-code counts).

\textbf{Implications and future work}: Poly-Encoders lay a framework for practical implementation in classrooms due to their accessibility on standard hardware and local-run privacy benefits. Future research should explore multidimensional assessment (novelty, feasibility, value) and generalize to other tasks such as the Alternative Uses Task (AUT). Additionally, incorporating textual justifications for model scores could further enhance accuracy and interpretability. To support reproducibility, all code and data are available online \footnote{https://github.com/sam-grouchnikov/ca-polyencoder-official}.

\textbf{Limitations}: This study was limited by the SCTT's dataset's label noise and the model's inability to generalize to unseen prompts. Potential biases include models preferring longer responses, and responses with writing styles from less common dialogues. BERT is pretrained on ``Standard English'' and thus lower scores may be assigned to responses that deviate from this. Such language may include African American Vernacular English (AAVE), regional dialects, or non-standard syntax. While fine-tuning BERT is more efficient than training LLMs, it still requires basic GPU access, which may limit some users. Most importantly, AI-based creativity assessment methods continue to function as ``black boxes.'' It remains difficult to fully interpret how and why such models assign specific creativity scores.  Enhancing model interpretability represents a crucial area for future investigation, as results without transparent reasoning provide an incomplete understanding of creativity evaluation.

\bibliographystyle{splncs04}
\bibliography{paperpile}

\end{document}